\documentclass[runningheads]{llncs}
\usepackage[T1]{fontenc}
\usepackage{graphicx}
\usepackage[misc]{ifsym}
\usepackage{url}
\usepackage[small]{caption}
\usepackage[colorlinks = true,
            linkcolor = blue,
            urlcolor  = blue,
            citecolor = black,
            anchorcolor = blue]{hyperref}
\usepackage{wrapfig}
\usepackage{sidecap}
\usepackage{color}

\begin{document}
\title{Cloud-Based Real-Time Molecular Screening Platform with MolFormer}
\toctitle{Cloud-Based Real-Time Molecular Screening Platform with MolFormer}

\author{
Brian Belgodere$^*$ \and
Vijil Chenthamarakshan$^{*}$ \and
Payel Das$^*$ \and
Pierre Dognin$^*$ \and
Toby Kurien$^*$ \and
Igor Melnyk$^*$ \and
Youssef Mroueh$^*$ \and
Inkit Padhi$^*$ \and
Mattia Rigotti$^*$ \and\\
Jarret Ross$^{*\scriptsize\textrm{\Letter}}$ \and
Yair Schiff$^*$ \and
Richard A. Young$^*$ 
}
\tocauthor{Brian~Belgodere, Vijil~Chenthamarakshan, Payel~Das, Pierre~Dognin, Toby~Kurien, Igor~Melnyk, Youssef~Mroueh, Inkit~Padhi, Mattia~Rigotti, Jarret~Ross, Yair~Schiff, Richard~A.~Young}
\authorrunning{Belgodere et al.}
\institute{IBM Research\footnote[0]{$^*$Equal contribution, ordered alphabetically. Contact author: rossja@us.ibm.com}}
\maketitle
\begin{abstract}
With the prospect of automating a number of chemical tasks with high fidelity, chemical language processing models are emerging at a rapid speed. Here, we present a cloud-based real-time platform that allows users to virtually screen molecules of interest. For this purpose, molecular embeddings inferred from a recently proposed large chemical language model, named MolFormer, are leveraged. The platform currently supports three tasks: nearest neighbor retrieval, chemical space visualization, and property prediction. Based on the functionalities of this platform and results obtained, we believe that such a platform can play a pivotal role in automating chemistry and chemical engineering research, as well as assist in drug discovery and material design tasks.
A demo of our platform is provided at \url{www.ibm.biz/molecular_demo}.

\end{abstract}
\section{Introduction}
Machine learning (ML) offers high throughput material exploration that is more efficient than high-cost quantum chemical/empirical force-field calculations and wet lab evaluations.
In this work, we present a cloud-based  platform for real-time virtual  screening of molecules, which uses a general-purpose deep learning model of large organic small molecule libraries.  %that is comprised of , prediction, and visualization. 
Specifically, our Molecular Explorer Platform  builds on our previous work “MolFormer”, a large, masked chemical language model trained on over 1.1 billion molecular string representations known as SMILES (see \cite{ross2021large} for details).
MolFormer provides representations for molecules that we showcase here in a platform enabling neighbor search, chemical space visualization, and property prediction for molecules of interest.

%\textcolor{red}{Payel}
%This short example shows a contrived example on how to format the authors' information for {\it IJCAI--22 Proceedings}.

\section{Real-Time Screening Platform}

\begin{figure}[ht!]
\centering 
  \includegraphics[width=0.9\linewidth]{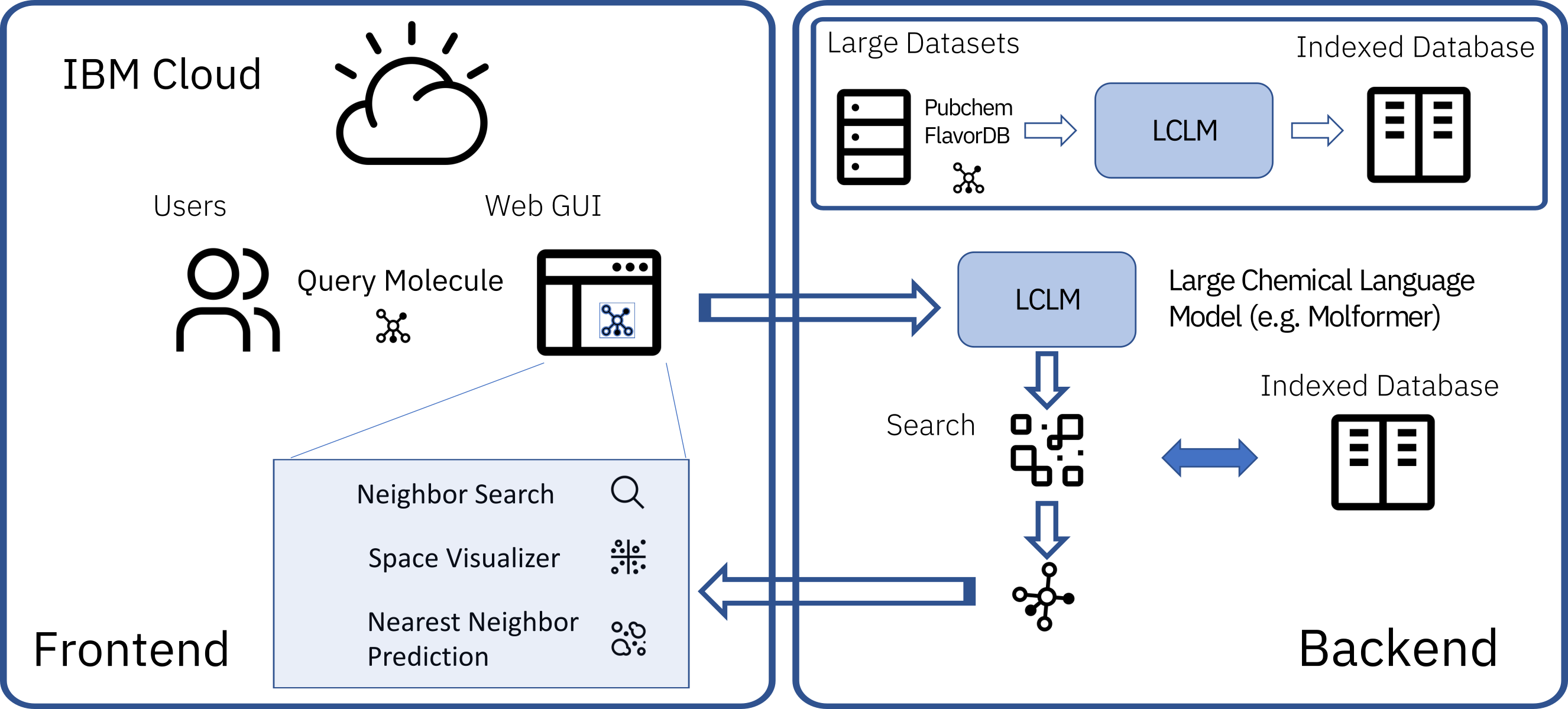}
  \caption{Diagram of our Molecular Explorer Platform.}
  \vspace{-0.5cm}
\end{figure}

Given a backend dataset, such as PubChem~\cite{pubchem/10.1093/nar/gkaa971} or FlavorDB \cite{flavordb}, we start by embedding this database through MolFormer and obtain a latent representation of 768 dimensions.
To index the database for nearest neighbor search, we start by reducing the dimensionality of MolFormer representations using Discrete Cosine Transform to 128 dimensions.
We then leverage the approximate nearest neighbor search library HNSWlib \cite{malkov2018}.
with hyperparameters calibrated so that retrieval would be faster than 10 milliseconds per query with a recall of 0.99.

Our molecular platform consists of a frontend GUI that enables 3 critical molecule screening functionalities: 1) neighbor search, 2) visualizing latent space of molecules using t-SNE visualization in 2D, and 3) nearest neighbor property prediction using Sklearn~\cite{scikit-learn} for moderately sized datasets and FAISS \cite{johnson2019} for large-scale predictions.
User queries are provided in the form of a line separated list or \texttt{.txt} file of molecule SMILES strings.%, with eventual labels separated by a comma.
An implementation of MolFormer running on OpenShift on IBM Cloud enables real-time feedforward embedding of SMILES strings, which are normalized using the RDKit library \cite{landrum2013rdkit,landrum2013rdkitsoftware}.
The obtained MolFormer representation is subsequently used to query the indexed backend database, which returns the user provided $N$ nearest neighbors along with molecular properties, such as logP, QED, and weight, which are computed on-the-fly using RDKit. Optional call to PubChem’s similarity search API is provided in our user interface allowing the user to compare it to  MolFormer similarity.%es to that of PubChem.
If a user provides property labels for each SMILES string, such as toxicity or flavor (see use case 2), the molecular platform enables visualization of the embedding space color-coded by labels in t-SNE 2-dimensional space.
Finally, nearest neighbor prediction functionalities using known properties of the backend index database are also provided, along with predictions for these properties of query molecules and graphical visualization of the results.%If provided with known property labels for query molecules, our molecular explorer platform returns a confusion matrix for nearest neighbor prediction.
\vskip -0.09in

%\subsection{Organization of the Explorer frontend (A) and backend (B) packages. }
%\subsection{front end/ back end/ infrastructure}
%\textcolor{red}{[Figure 1] Diagram flowchart [Pierre] \url{https://jcheminf.biomedcentral.com/articles/10.1186/s13321-021-00550-y}}
%\paragraph{MOLFORMER}
%\paragraph{Real-time/ Cloud based}
%\section{UI and functionalities  }

%\paragraph{Nearest Neighbor Search}
%\paragraph{Space Visualizer}
%\paragraph{Nearest Neighbor Prediction }

\begin{figure*}[ht!]
\centering 
  \includegraphics[width=0.85\linewidth]{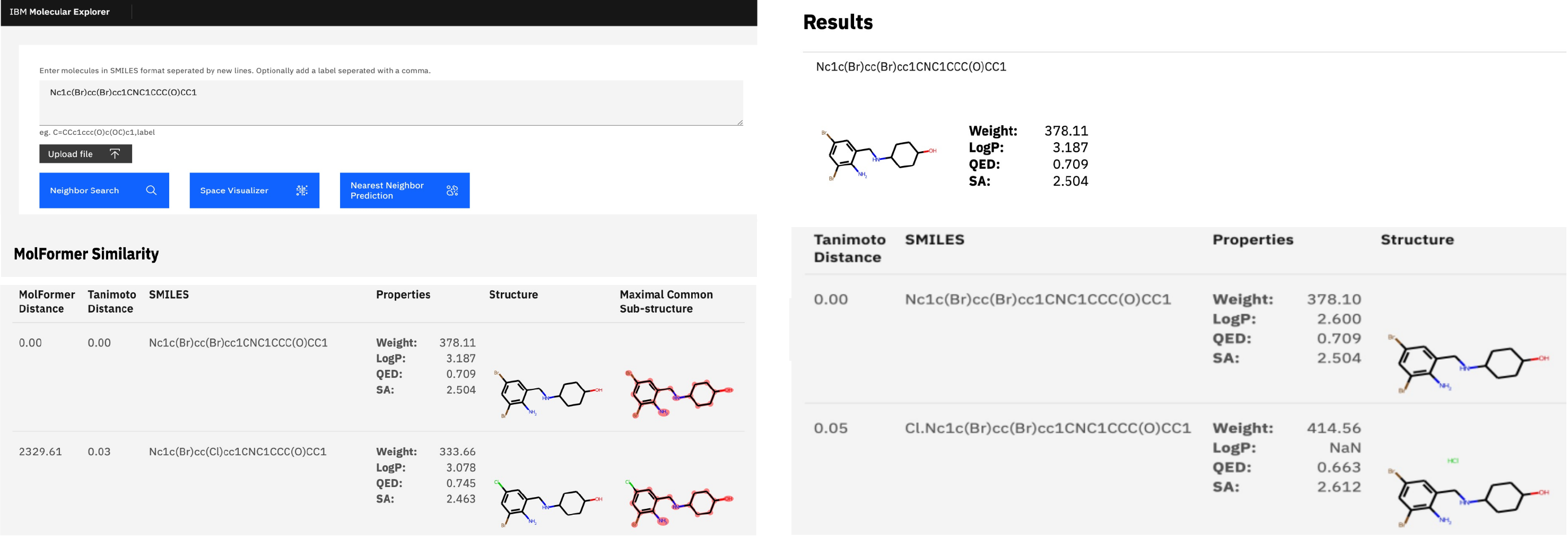}
  \caption{Use Case 1: Nearest neighbor search in large chemical embedding space.}
  \label{covidusecase}
  \vspace{-0.7cm}
\end{figure*}

\subsubsection{Use Case 1: Similarity search among known drug molecules.}

A typical task that arises in molecule screening/discovery is to identify similar molecules in existing chemical libraries.
This is a frequent use case for medicinal chemists, for example.
Our molecule explorer platform allows users to retrieve similar molecules from PubChem using the PubChem API~\cite{pubchem/10.1093/nar/gkaa971} and MolFormer embeddings. To achieve this, we index over 100 Million molecules from pubchem  embedded in the MolFormer latent space. 
As an example, we show the neighbor retrieval results for  known drug molecules (Table S4 from ~\cite{Hoffman_Chenthamarakshan_Wadhawan_Chen_Das_2022}) obtained using the platform in Figure~\ref{covidusecase}. The maximal common subgraph of the query molecule and closest molecules are also shown allowing a user to understand the key differences between the query molecules and its closest neighbors.

\subsubsection{Use Case 2: Flavor molecules screening.}

\begin{wrapfigure}{r}{0.42\textwidth}
 \centering
  \vspace{-0.95cm}
  \includegraphics[width=\linewidth]{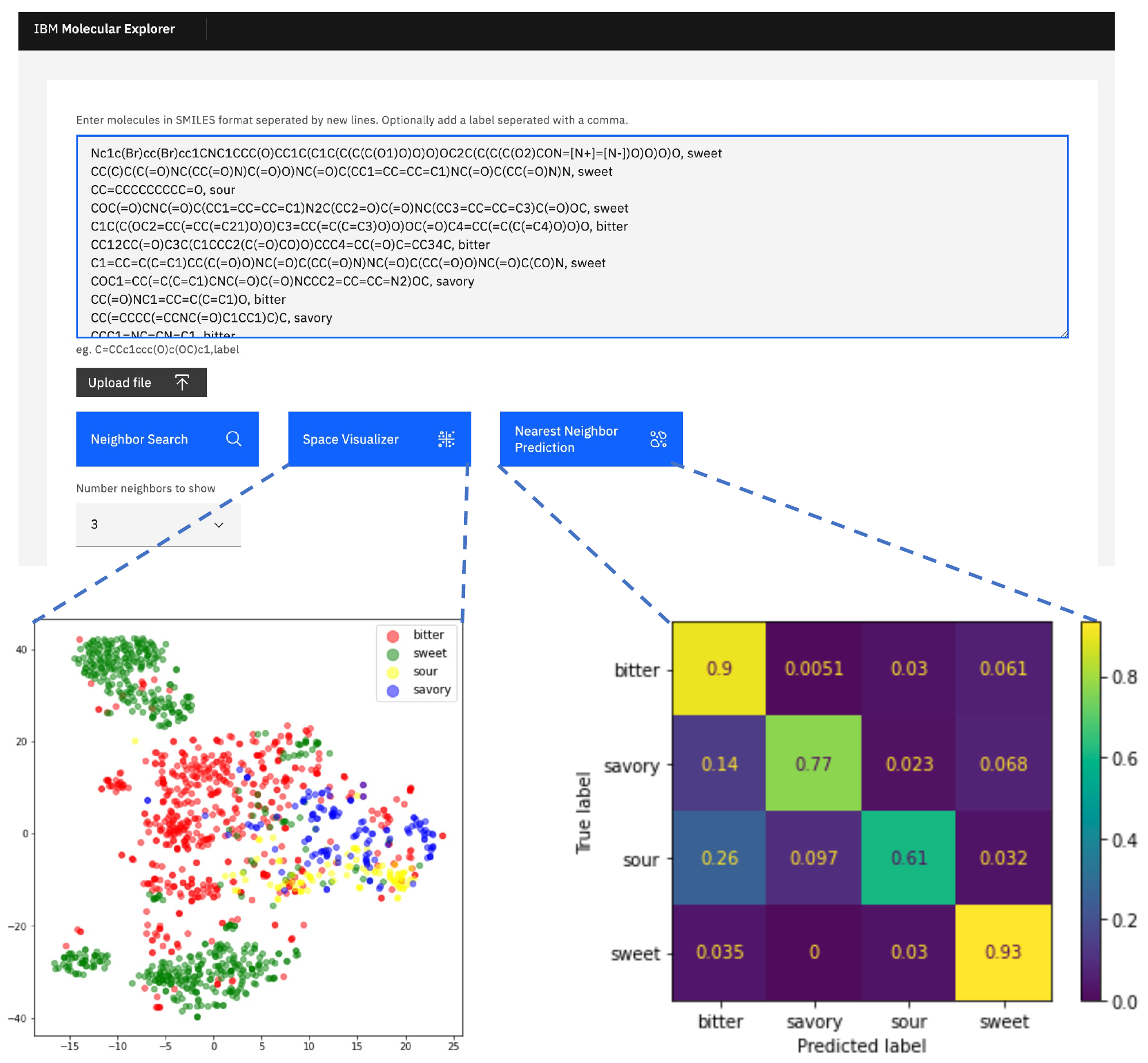}
  \caption{Use Case 2: Visualization of unsupervised  MolFormer Embeddings in t-SNE space and separation of flavor molecules in that space.}
  \vspace{-0.7cm}
  \label{usecase2}
\end{wrapfigure}

The molecule explorer platform also allows a user to upload a set of molecules along with a their corresponding class (property) labels and visualize their chemical space.
The user can visually explore the t-SNE \cite{JMLR:v9:vandermaaten08atsne} representation of those molecules obtained using   %and evaluate whether
MolFormer embeddings and check if the resulting chemical space captures the distribution of class labels for a particular application.
Alternatively, a k-NN classifier can be trained on the MolFormer embeddings and performance characteristics of the classifier can be visualized as a confusion matrix.
We show the application of these techniques to molecules with different flavor descriptions from \cite{flavordb}.
The flavor database consists of 25,595 individual flavor molecules with up to 43 different attributes.
%Each molecule consists of up to 43 different attributes including SMILES representation, molecular weight, charge, odor properties, flavor profile, and a synthetic/natural determination for most molecules.
4 basic flavors were chosen for evaluation; bitter, sweet, sour, and savory.
%The bitter and sweet classes consist of 600 individual molecules each while the sour and savory molecules only contain 98 and 133 molecules, respectively.
Figure~\ref{usecase2} shows that our chemical space map captures the different flavors and provides excellent predictive performance.

%We further predict xlogp and toplogical polar surface area using a k-NN Regressor achieving an RMSE of 1.13 and 25.94 respectively.

%\textcolor{red}{t-SNE PLOT +  confusion matrix knn Vijil} 

%flavor (upstream) 
%Payel: questions - (1) can we add tasteless/blant as the fifth class? (2) chem properties for k-NN analysis - xlogP, charge, volume3D, topological/ polor/surfacearea. (3) charge is known to be important for flavor prediction - can just color code t-SNE plot with corresponding charges as a first try (edited) 

%+ chemical (downstream) just do knn on it  [Dataset of all four flavors , labels] 

\subsubsection{Use Case 3: Drug-like molecules screening.}
\begin{figure*}[ht!]
\centering 
  \includegraphics[width=0.55\linewidth]{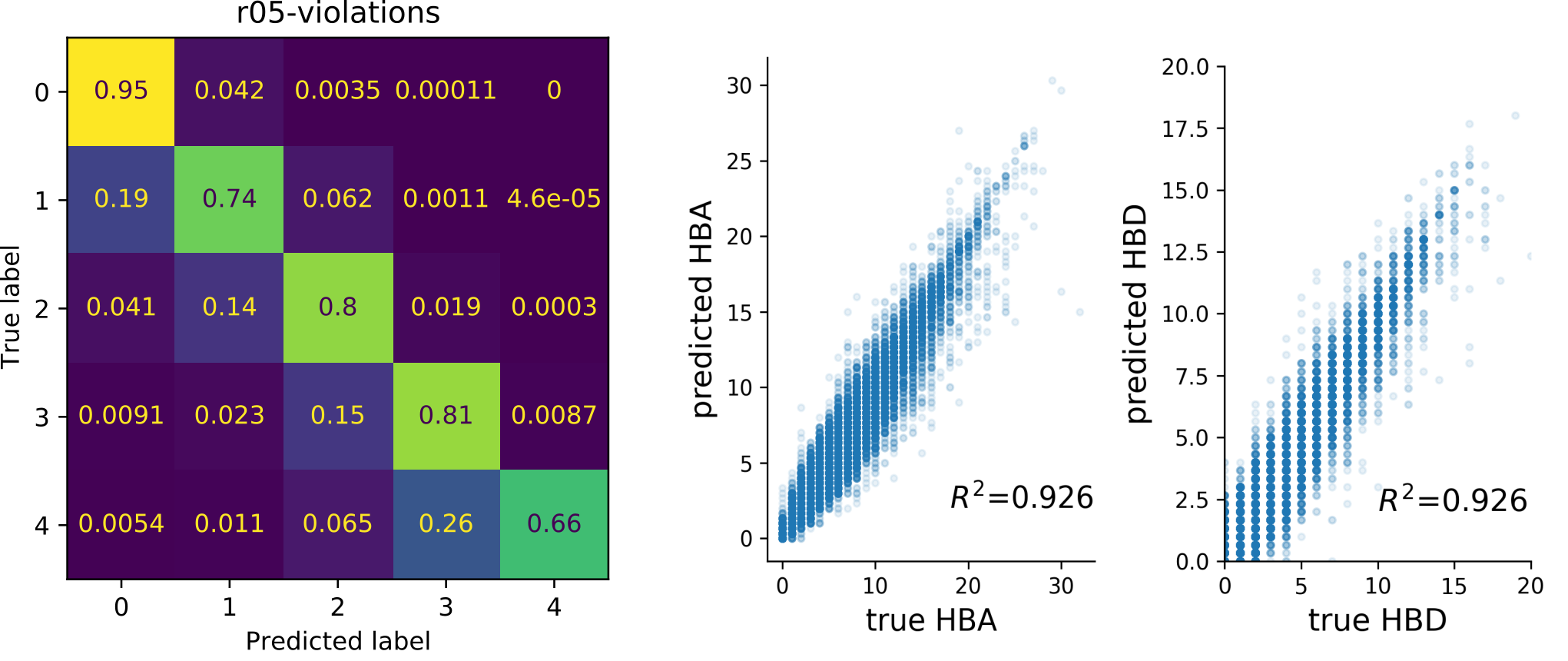}
  \caption{Use Case 3: Retrieval of r05-violations of 1.8M drug-like molecules with 1-NN gives an average holdout prediction accuracy of 0.90 (left). HBA and HBD are also predicted with high accuracy ($R^2=0.926$ for both) by $k=3$ NN-Regression (right).}
  \label{usecase3}
  \vspace{-0.4cm}
\end{figure*}

%\textcolor{red}{Mattia, Payel}
% Mattia/ Jerret knn acccuracy  
Lastly, we predict the conformity to the RO5 (Lipinski rule of five) of 1.8M molecules out of $\sim$2M from the CheMBL dataset \cite{ChemBL}, which presented SMILES representations.
A k-NN classifier was trained on 1.44M MolFormer embeddings with the FAISS library \cite{johnson2019} and used to predict RO5 violations of 360k held-out molecules based on their neighbors, resulting in a classification accuracy of 90\% (see Fig.\ \ref{usecase3}).
%We selected 360k molecules at random as a holdout dataset and predicted their RO5 violations based on the number of RO5 violations of their nearest neighbor in MolFormer embedding space among the remaining 1.44M ``training'' molecules.
%This could be done efficiently thanks to the FAISS library \cite{johnson2019} and resulted in an accuracy of 0.90 on the 5-classes classification task of predicting the number of RO5 violations (see confusion matrix in Fig.\ \ref{usecase3}).
We then predicted HBA (hydrogen bond acceptor) and HBD (hydrogen bond donor) on the same split by averaging the HBA and HBD values of $k=3$ nearest neighbors, obtaining high coefficients of determination of $R^2=0.926$ (see Fig.\ \ref{usecase3}).

\bibliographystyle{splncs04}
\bibliography{main.bib}
\end{document}